\documentclass[runningheads]{llncs}
\usepackage[T1]{fontenc}
\usepackage{graphicx}
\usepackage{hyperref}
\usepackage{color}

\usepackage[linesnumbered,ruled,vlined]{algorithm2e}
\usepackage{epsfig}
\usepackage{subcaption}
\usepackage{calc}
\usepackage{amssymb}
\usepackage{amstext}
\usepackage{amsmath}
\usepackage{multicol}
\usepackage{pslatex}
\usepackage{enumitem}
\usepackage[bottom]{footmisc}
\usepackage{xcolor}
\usepackage[final]{pdfpages}
\usepackage{xspace}
\usepackage{lmodern}
\usepackage[T1]{fontenc}
\usepackage{textcomp}
\usepackage{etoolbox}
\usepackage[utf8]{inputenc}
\usepackage{stmaryrd} 
\newcommand{\xn}{\noindent}
\newcommand{\xv}{\vspace{0.1cm}}

\newcommand{\xxy}{(x,y)}

\newcommand{\xbrk}{\linebreak[3]}

\begin{document}
\title{Meta-autoencoders: An approach to discovery and representation of relationships between dynamically evolving classes\\ 
\vspace{0.2cm}
{\small A work-in-progress report}}
\titlerunning{Meta-autoencoders}
%
\author{Assaf Marron\inst{1}\orcidID{0000-0001-5904-5105} \and
 Smadar Szekely\inst{1}\orcidID{0000-0003-1361-1575} \and \\ 
Irun Cohen\inst{2}\orcidID{0000-0002-3906-6993} \and David Harel\inst{1}\orcidID{0000-0001-7240-3931}} 
\authorrunning{A. Marron et al.}
%
\institute{Dept. of Computer Science and Applied Mathematics, 
\and 
Department of Immunology and Regenerative Biology, \\ Weizmann Institute of Science,  Rehovot, 7610001, Israel \\ 
\email{<firstname>.<lastname>@weizmann.ac.il}
}
\vspace{-1.1cm}
\maketitle              
\begin{abstract}
An autoencoder (AE) is a neural network that, using self-supervised training, learns a succinct parameterized representation, and a corresponding encoding and decoding process, for all instances in a given class. 
Here, we introduce the concept of a \emph{meta-autoencoder} (MAE): an AE for a collection of autoencoders. 
Given a family of classes that differ from each other by the values of some parameters, and a trained AE for each class, an MAE for the family is a neural net that has learned a compact representation and associated encoder and decoder for the class-specific AEs. One application of this general concept is in research and modeling of natural evolution --- capturing the defining and the distinguishing properties across multiple species that are dynamically evolving from each other and from common ancestors. In this interim report we provide a constructive definition of MAEs, initial examples, and the motivating research directions in machine learning and biology. 

\end{abstract}
\vspace{-0.8cm}
%
%

\section{Introduction}\label{sec:intro}

\begin{figure*}[h] 
	\centering
    \hrule
	\includegraphics[width=1.0\linewidth]{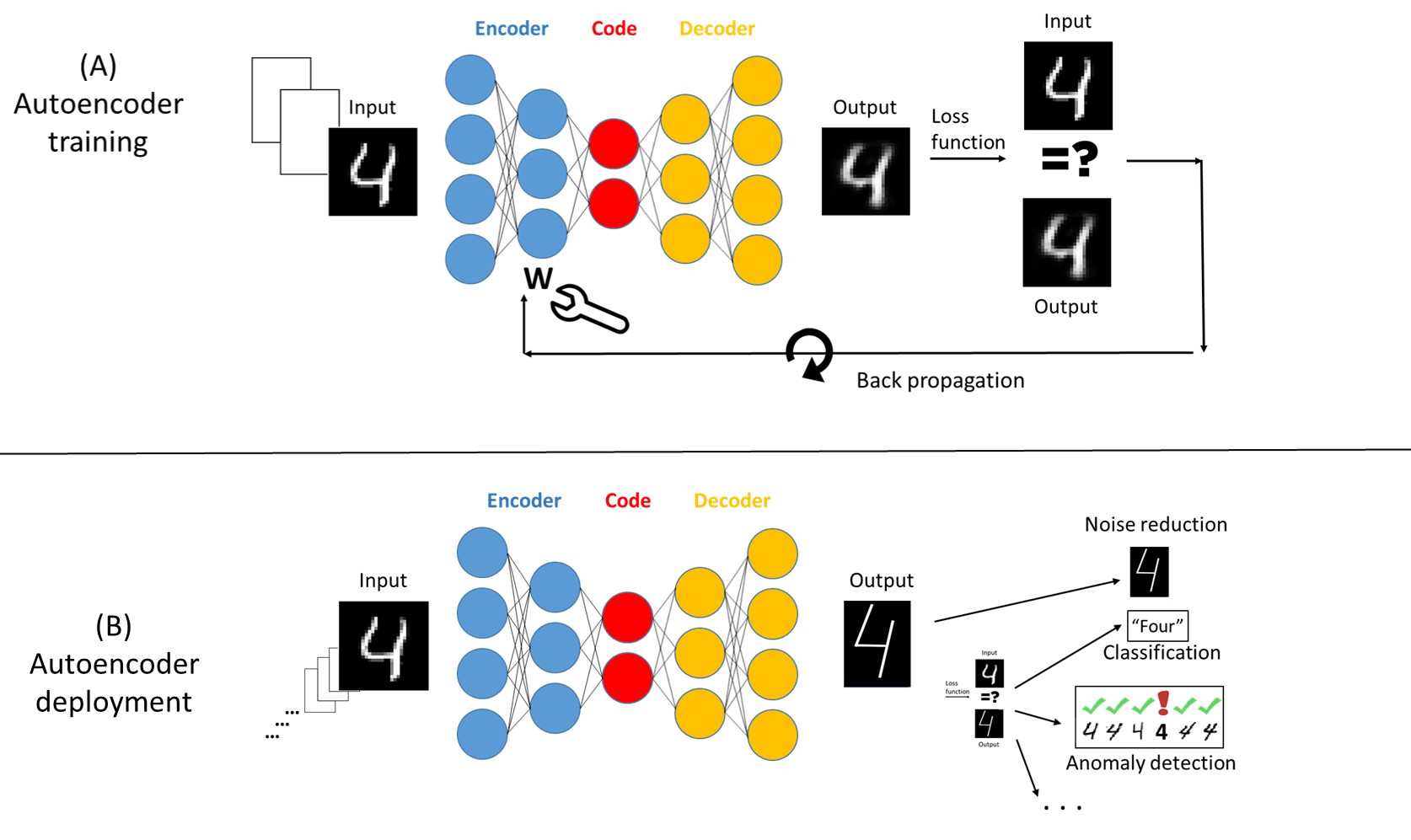}
	\caption{Recap of artificial autoencoding. 
	 A typical AE is a NN consisting of an encoder (blue circles), a code layer, also called the bottleneck or latent layer (red circles), and a decoder (orange circles). Inputs (here, images of handwritten digits) are encoded by the encoder into the latent layer, and then reconstructed by the decoder. Training (A) with a finite set of examples, the differences between the inputs and the reconstructed outputs are computed, and the weights W and the biases (not shown) are adjusted in backpropagation and gradient descent to minimize the reconstruction loss. 
Once trained, the AE is deployed (B) to perform its inference or application task  (noise reduction, classification, anomaly detection, etc.) using the now fixed encoding and decoding computations processes on an unbounded number of inputs from the domain of interest. 
}
  \hrule
	\label{fig:artificialAutoencoding}
	
\end{figure*}

\begin{figure}[h] 
\hrule
	\centering
	\includegraphics[width=0.5\linewidth]{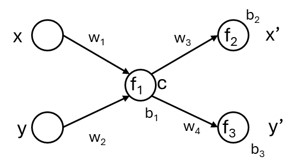}]
	\caption{Autoencoder example. This  AE has 3 layers (with 2, 1, and 2 neurons respectively), weights $w_1,w_2,w_3,w_4$,  biases $b_1,b_2,b_3$,  and activation functions $f_1, f_2,f_3$. For every input $(x,y)$,  $c = f_1(x w_1+y w_2)+b_1$;  $x' = f_2(c w_3)+b_2$, and $y' = f_3(c w_4)+b_3$. In a perfectly trained AE, $x=x'$ and $y=y'$.}
\vspace{0.2cm}
\hrule
\label{fig:autoencoder212}
\end{figure}

An \textit{autoencoder} (AE) for a class of instances is a neural network (NN) that, following self-supervised training, establishes a dimensionality-reducing representation for the class. 
~\cite{GoodfellowBengioCourville2016DeepLearningBook,berahmand2024autoencodersReview}. Fig.~\ref{fig:artificialAutoencoding} (borrowed and adapted from~\cite{cohenMarron2023autoencoding}) provides a brief introduction to autoencoding, and Fig.~\ref{fig:autoencoder212} depicts an example AE in more detail. 
For illustration, consider the set of all circles in the plane with center at $c=(0,0)$. 
For each radius $r$, the circle $C_r$ is a class containing all the points in on that circle; together, these classes form a family $F$ of classes, distinguished from each other by the value of $r$. 
For a given $r$, any point $p=(x,y)$ in $C_r$ can be represented by a single number, for example, the angle between the radius to $p$ and the positive $x$ axis; one can train an AE that 
computes such encoding and its associated decoding. 
A meta-autoencoder (MAE) for $F$ is an AE that can succinctly encode and decode AEs of the individual classes. A formal constructive definition appears in Section~\ref{sec:defs}. 
Given the defining information for an AE for some circle, the MAE's encoder would encode this information into a succinct code, perhaps just the radius, and the MAE's decoder would produce all the information needed to construct a valid AE for that circle. The reconstructed AE may or may not be identical to the original input AE. 

Our motivation for introducing and investigating MAEs is anchored
in our research of the evolutionary theory of Survival of the Fitted
~\cite{cohenMarron2023autoencoding,cohenMarron2020survivalOfTheFitted,marronSzekelyCohenHarel2025sexualReproduction}.
According to this theory, the biosphere is sustained mainly by the dynamics of its  interaction networks. In a process that we term \emph{natural autoencoding}, 
iterative imperfect reproduction and differential survival drive the emergence, reshaping and sustainment of interaction networks that external observers may label as organisms, species or ecosystems. 
Clearly, natural autoencoding differs from artificial autoencoding in various aspects, most conspicuously the absence of loss functions and backpropagation. Our current plans for modeling natural autoencoding include AEs and MAEs with modifications, constraints and a controlled execution environment.
A bird's eye view of one such setup includes iterations of: (i) using AEs  as imperfect reproduction engines for a population; (ii) clustering of the individuals in new generations; (iv) dynamic interaction and differential survival among individuals and groups; and (v) training and retraining of AEs and MAEs for the classes comprising the surviving population. 

In this model, the emergent traits that are characteristic and essential to each dynamically created cluster, i.e., species, are not predictable, and neither are patterns of  differences between such species. 
Iterative training of AEs and MAEs can help in discovering and representing such features. 
We find this modeling approach attractive as it offers common principles across different domains, 
uses computational elements that have been shown to exist in nature, and readily accommodates specialized low level components, as may evolve in nature for sensors and actuators, and may be used in the context of neural nets in feature engineering and in activation functions. This is contrasted with machine learning approaches, where a highly sophisticated network is designed and constructed in order to tackle a difficult challenge. 

In Section~\ref{sec:defs} we first formalize the concept of MAEs, and discuss various aspects of the processes for creating them. 
In Section~\ref{sec:examples} we exemplify the construction of MAEs. In Section~\ref{sec:relwork} we position this direction with regard to related work in statistics and in machine learning. 
In Section~\ref{sec:conclusion} we outline
our motivating research trajectories for the theoretical concept of MAEs and for the specific application of AEs and MAEs in modeling evolution in multi-species ecosystems. 

\section{Meta Autoencoders}\label{sec:defs}

\xv \xn \textbf{About scope restriction.} The general concept of an MAE is broader than is needed in the present introduction and initial analysis. The definition below is for a restricted case; still, here, we keep the terms MAE and others without constraining adjectives. 

\xv \xn \textbf{Domain.}
We constrain the discussion below to $R^2$, the Cartesian plane, where every  instance $z$ can be represented with the coordinate pair $z=\xxy$.

\xv \xn \textbf{Encodable Classes.} 
We say that a class $C$ of instances in $R^2$ is \emph{encodable to the domain $T \subseteq R$, within $\epsilon$, for $\epsilon \in R$} (or just \emph{encodable}, for short),  if there exists at least one pair of encoding and decoding functions 
$f_e^C : C \rightarrow T~~$ and $f_d^C : T \rightarrow R^2~~$~ 
respectively, such that 
for all $z \in C$: 
$$\|z - f_d^C(f_e^C(z))\| \leq \epsilon ~~ .$$  
In other words, there exist a function that can encode each point $z=\xxy$ in $C$ into a single real number $t \in T$, and another function that can decode that $t$ into another point $z' \in R^2$, whose distance from $z$ is small enough for the purposes at hand. Note that $z'$ is not necessarily in $C$.

For example, the class $C_a$ of points on the straight line $y=a \cdot x$ is encodable into $R$ for an arbitrarily small $\epsilon$. Examples of pairs of encoding and decoding functions for this class can be $f_e(x,y) = x$ \xbrk and $f_d(t)=(t,a t)$, respectively, or  $f'_e(x,y) = 2 \cdot x$ \xbrk and $f'_d(t)=(\frac{t}{2},a \frac{t}{2})$. In another example,  the class $C_r$ of points on the circle of radius $r$ and centered at the origin can be encoded by \xbrk $f_e(x,y)=\operatorname{atan2}(y,x)$  \xbrk and decoded by \xbrk $f_d(t)=(r \cos(t), r \sin(t))$. 

\xv \xn \textbf{Autoencodable classes.} 
We say that an encodable class $C$ is \emph{autoencodable}, if there is a known AE for it. This is not about the theoretical existence of an AE for $C$, but about whether one was already created, or a way was created to construct one at will. Efforts to create an AE for a given class $C$ may fail because the class is not encodable, or, due to hurdles in the AE training process.  One such hurdle is manifested in encoding functions that are too complex for standard AE architectures, even with preparatory feature engineering steps where inputs are pre-processed before being fed to the NN. 

\xv 

\xv 
\xv \xn \textbf{MAE: A constructive definition.} 
Let $F$ be a family of autoencodable classes. We say that a given neural network $M$ is an MAE for $F$, if $M$ is an AE that was successfully constructed as follows:  (i) an arbitrary or random sample $F^0$ of classes in $F$ was selected;  
(ii) a single AE architecture was selected (number of layers, number of neurons in each layer, connectivity, and activation functions); 
(iii) for each class $C \in F_0$ one or more AEs were created with the selected architecture, yielding $F^0_{AE}$, a set of AEs; (iv) the AE $M$ was successfully trained and tested on $F^0_{AE}$~ $\qed$. 

\xv 

\xv 
The conditions for calling $M$ an MAE for $F$ are quite relaxed. 
Thus, once constructed in this manner, $M$ may not be able to successfully encode and decode  all AEs (of the selected architecture) for all classes $C \in F$, or decode every synthetic code in the latent domain into a valid AE for some class in $F$. 
Furthermore, for a given family of autoencodable classes, an MAE may not exist, or the construction of one may be impractical.

\xv \xn \textbf{Feature engineering in MAE construction.}\label{sec:MAEArch} 
When choosing the class family for meta-autoencoding, one may already have in mind the parameters that characterize the individual AEs. However, this characterization may not be readily computable by standard NN architectures from the weights and biases of the AEs, calling either for richer activation functions or for feature engineering. As our focus is on biological modeling, we find these two options acceptable. as they align with our assumptions that natural learning processes have common and generic principles, and that biological mechanisms that are complex yet well-encapsulated and internally modular can evolve naturally. 

\begin{figure}[h] 

	\centering
	\includegraphics[width=1.0\linewidth]{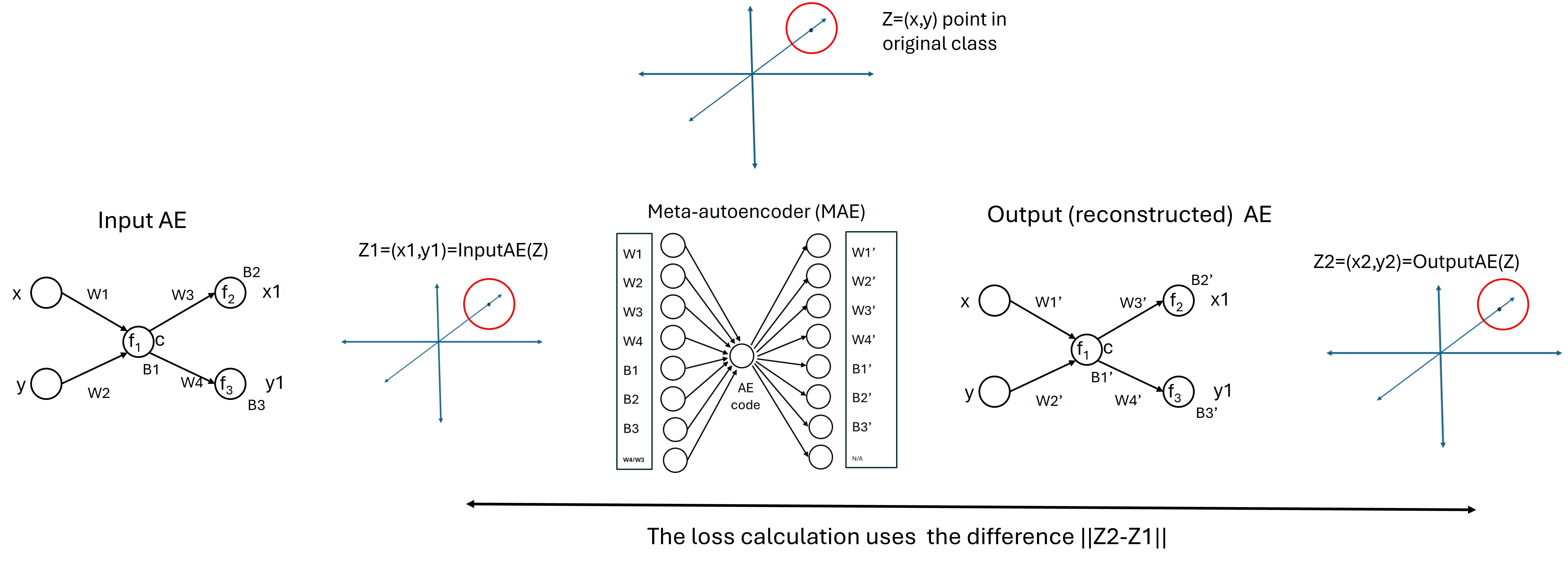}
	\caption{Execution-Driven Loss Calculation. When training the MAE, the MAE encodes an Input AE,  reconstructing it as a corresponding Output AE. A sample point Z from the class associated with the Input AE is encoded and decoded by both the input and output AEs, yielding  Z1 and Z2. The difference ||Z2-Z1|| ,  is used in loss calculation over a sample of several points.}
\vspace{0.2cm}
\hrule
\label{fig:execDrivLossPoal}
\end{figure}

\begin{figure}[h] 
 \vspace{0.2cm}
\centering
\includegraphics[width=1\linewidth]{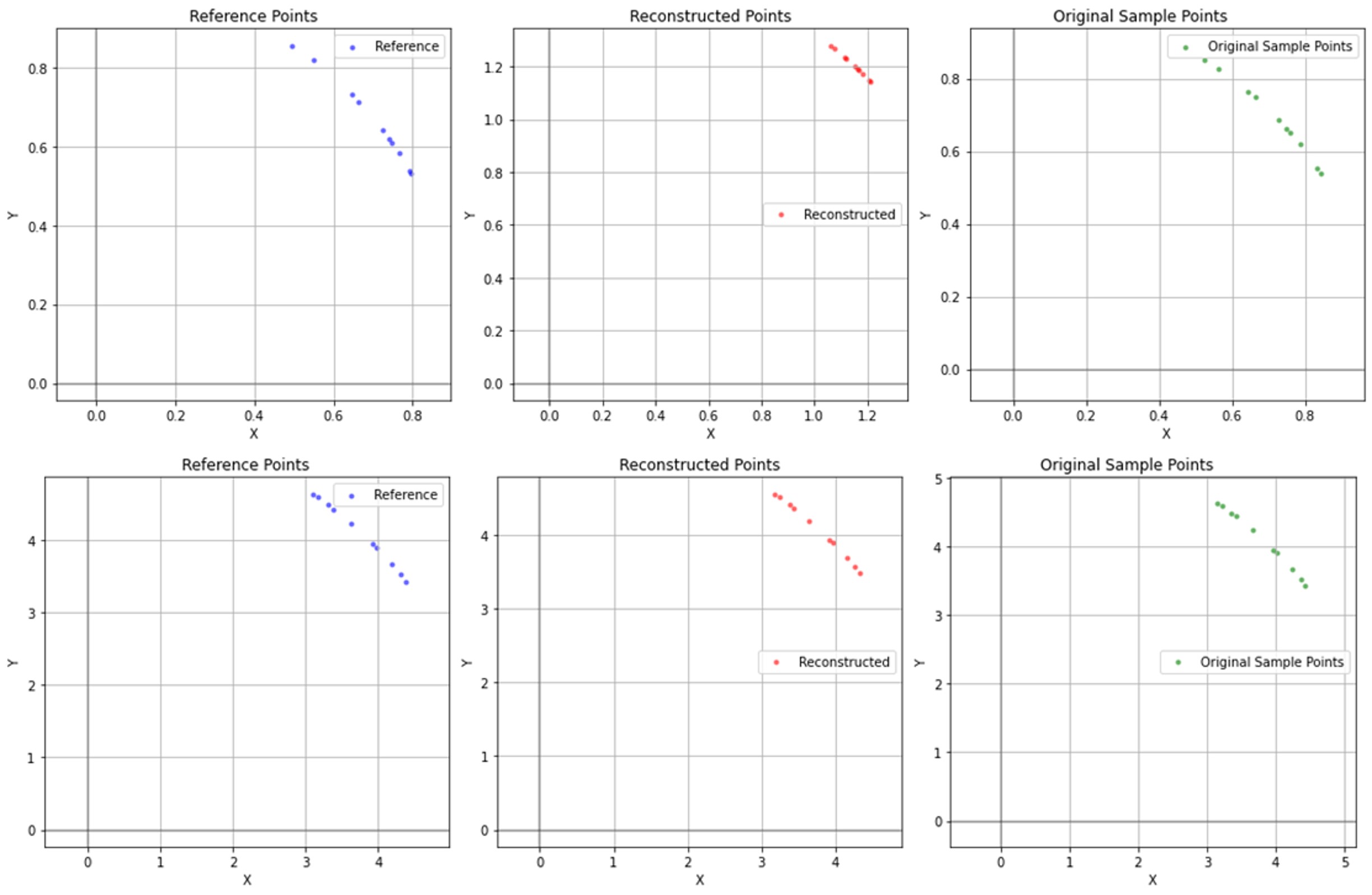}
\caption{Sample data for execution-driven loss calculation in one of the batches of AEs, in one of the epochs of MAE training for the family of Points-on-Arc classes.
While Fig.~\ref{fig:execDrivLossPoal} concentrates on a single point in a single AE, here each row of three images is for a separate AE, and shows multiple points for that AE.  The right image shows the original points in the class (an arc with a particular radius);  the left image shows the points reconstructed by the trained AE, and are thus very close to the original ones; the center image shows the points reconstructed by the output AE, as reconstructed from the trained AE by the MAE being trained. In the top row, the reconstructed AE preserves the arc shape, but the density of the points and the arc radius differ from the ones produced by the input AE. In the bottom row, the reconstructed AE has better performance. The loss calculation spans a batch of such input AEs, with multiple sample points for each. Note also that the final loss can vary between different AEs.} 
 \vspace{0.2cm}
 \hrule
 \label{fig:execDrivLossPoA}
\end{figure}

\xv \xn \textbf{Execution-driven Loss Calculations.}\label{sec:lossCalc} In NN training, classical loss calculations involve comparing the network's output with a desired output. When the network's output is a specification of a program or a process, classical loss functions compare the output specification to a desired specification. However, in various contexts, the loss calculation involves compiling and executing the program specification, and comparing the output of that execution to a predefined collection of desired outcomes~\cite{ye2022executionBasedBackpropagation}. 
While not required in the definition of MAEs, in our experiments, we extended such execution-driven loss calculation approach as follows.
In training, each input $AE$ of the MAE is an autoencoder that was constructed by training over a sample set of points drawn from a given class $C$. The reconstruction process carried out by the MAE yields an output autoencoder $AE'$. 
To calculate the loss, we sample a set of points $C^0 \subset C$ , and for each point $\xxy \in C^0$ we compute the distance $$\|AE'\xxy - AE\xxy\|~~.$$  Illustrations of execution-driven loss calculation appear in Figs.~\ref{fig:execDrivLossPoal} and~\ref{fig:execDrivLossPoA}.

Comparing the functionality of parent and child programs aligns with theories of evolution where sustainability depends on the ability of offspring to function in the same interaction networks that their parents' had participated in~\cite{marronSzekelyCohenHarel2025sexualReproduction}. 

\textbf{Note.} Within a common ML platform like Tensorflow, testing a reconstructed AE while training the MAE requires configuring the invoked AE as a stateless NN model, to avoid data structure collisions.  

\section{Meta-autoencoder Examples}\label{sec:examples}
Below we report our experience with the construction of MAEs for two classes of points in $R^2$: points on a line (PoaL) and points on a circle (PoaC). 

\subsection{Points on a Line (PoaL)}\label{sec:PoaL}
\begin{figure}[h] 
\vspace{0.2cm}
	\centering
	\includegraphics[width=0.7\linewidth]{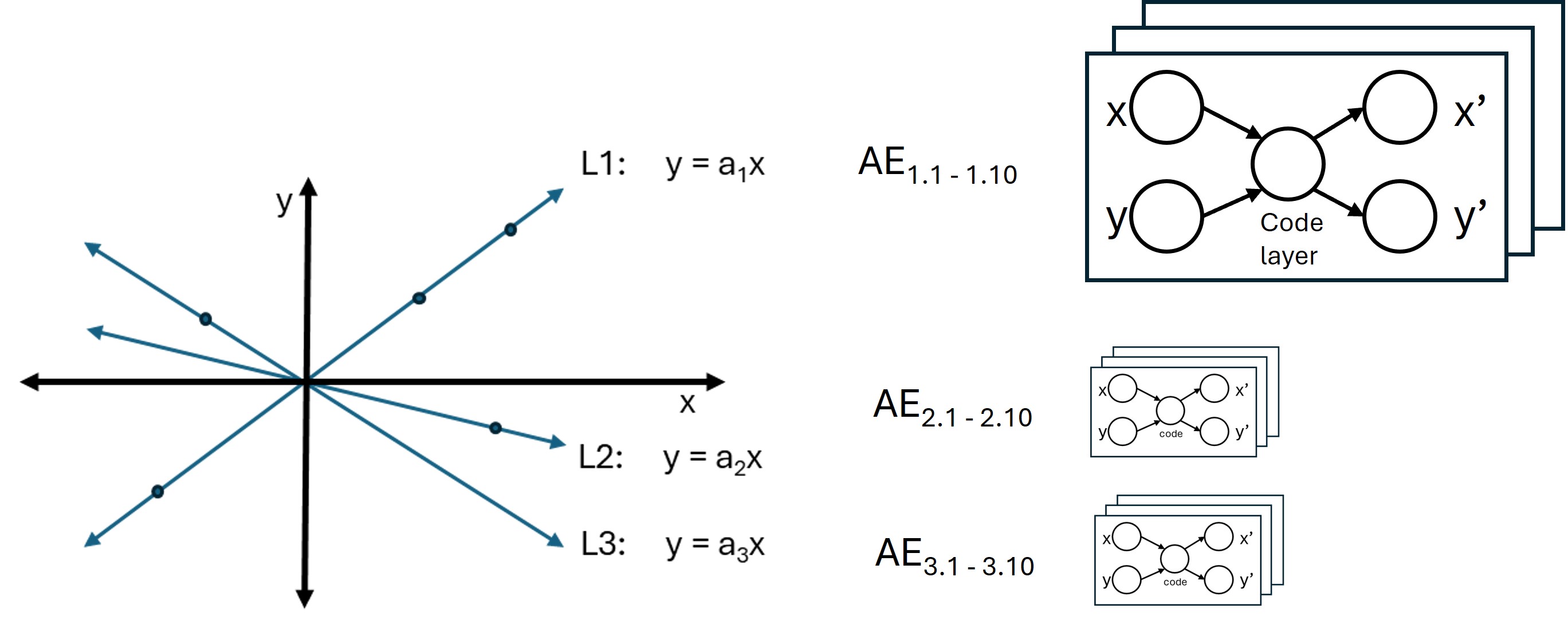}
	\caption{ Preparing the training set for Points-on-a-Line MAE. We sample classes ($L1$, $L2$, $L3$) in the family $F$  of straight lines in $R^2$ that go through the origin. The lines have slopes $a_1,a_2,a_3$, respectively. The elements of each class are all the points of the respective line. For each class we train 10 AEs, using a different set of 1000 random points for each.}
\vspace{0.2cm}
\hrule
\label{fig:PoalTrainingSet}
\end{figure}

Our first set of experiments is with family $F$ of classes $C_a$ where each class is the collection of all points on straight line  in $R^2$ 
that passes through the origin $(0,0)$, and whose angle $\theta$ with the $x$-axis, specified in degrees,  is an integer, and satisfies $-80 \leq \theta < 80$ (see Fig.~\ref{fig:PoalTrainingSet}. Each class is uniquely associated with the equation $y=ax$ where the line slope $a$ satisfies $a=\tan(\theta)$.  Every such class is autoencodable; one such example AE has: 
(i) three layers in a 2-1-2 arrangement as in Fig.~\ref{fig:autoencoder212}; (ii) weights of $w_1=1$ and $w_2 = 0$ for the encoder edges, and  $w_3=1$ and $w_4=a$ for the decoder edges; (iii) all biases are zero; and (iv) all activation functions are the identity, $f(x)=x$. 

For training the PoaL MAE, we created a training set of  PoaL AEs 
as follows. We kept the above 2-1-2 AE configuration with identity activations. 
We created 10 AEs for each of the 160 lines in the range for a total of 1600 AEs. 
Each of the 10 AEs of a given line was trained with a different set of 1000 random points on that line. 
We did not force the biases to be zero. 
See Fig.~\ref{fig:PoalTrainingSet}
The test set for each line AE was yet another set of random points. 

We constructed an MAE for the PoaL AEs as follows.
Given the fixed 2-1-2 architecture and the identity activations, each AE's NN has 7 defining parameters: 4 weights and 3 biases. 
Experiments with a na\"ive 7-1-7 architecture for the MAE did not readily provide good results. 
As stated above, one of the codes that represents such an AE  and that consists of a single real number, could be the value of $a$, which in this case satisfies $a=\frac{w_4}{w_3}$. Thus, at least for one of the options of the code computations, 
an MAE may have to calculate or approximate the above ratio. However, computing division in a neural net may require using logarithms in activation functions, or two-argument functions, which are non standard. We chose to solve the problem with feature engineering, and provided $\frac{w_4}{w_3}$ as one of the inputs, yielding an 8-1-8 neural net.

\begin{figure}[h] 
\hrule
\vspace{0.2cm}
	\centering
	\includegraphics[width=0.6\linewidth]{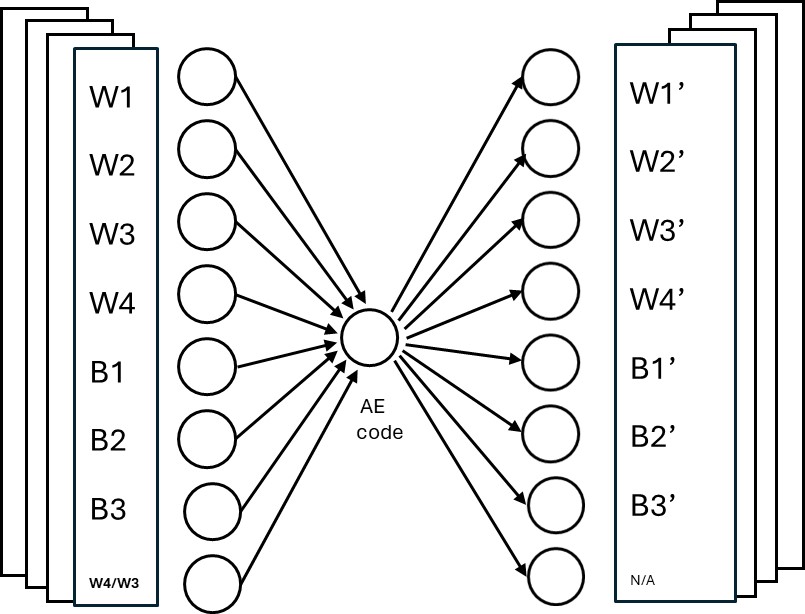}
	\caption{MAE for Points-on-a-Line AE. Each input AE is defined by its 4 weights and 3 biases. An added 8th feature is the ratio W3/W4. The MAE NN outputs new weights and biases, to be used in a reconstructed AE.
    }
\vspace{0.2cm}
\hrule
\label{fig:MAE818}
\end{figure}

The architecture of the PoaL MAE  is shown in Fig.~\ref{fig:MAE818}. 

The training process used execution-driven loss calculation as described in Section~\ref{sec:defs}. Specifically, for the input AE with a slope $a$ ($=\tan(theta)$ for some angle $\theta$), we picked 4 points  on that line: $(-10,-10a)$,$(-3.33,-3.33a)$,$(3.33,3.33a)$, and $(10,-10a)$, and compared the outputs of the input AE and of the reconstructed AE, when processing these 4 inputs.  
The learning converged well, yielding a working meta auto-autoencoder, that could encode and decode any newly trained AE for any line in the above $F$. The details of the  resulting 8-1-8 MAE NN are provided in (URL to be provided in CR version). 

This experiment can be extended in various ways, including: (i) allow $F$ to contain classes in  a continuous range of the angle $\theta$; (ii) allow vertical lines; (iii) replace  feature-engineering heuristics with additional NN layers; (iv) feed a broader range of computed features to the training process; and, (v) interpret the learned AE and MAE codes. 

\subsection{Points on a Circle (PoaC)}\label{sec:PoaC}

In this example $F$ is the family of classes where each class is the collection of all points on a circle  in $R^2$, whose center is at (0,0). 
As stated earlier, given a class $C_r$ with its radius $r$, each point $\xxy \in C_r$ can be encoded as the angle formed between the radius to the point and the positive $x$-axis, which satisfies $\theta = \operatorname{atan2}(y,x)$. 
Ideally and AE for $C_r$ would compute this function, however, a standard NN can only compute an approximation of this function.
We trained each class AE as follows. We used an architecture of 5 layers having 2, 8, 1, 2, and 2 neurons, respectively, expecting layer 3 to capture the angle approximation, as encoded in layers 1-3  
with $\operatorname{tanh()}$ activations. 

In the decoder we applied heuristics:
in layer 4 we used activation of $\cos()$ and  $\sin()$, 
and fixed the two weights from the code neuron in layer 3 at 1.   
In layer 5 we used identity activations, 
and assumed that each of the two neurons of layer 4 is connected to just one respective neuron in layer 5. We  expect, but do not force,  the trained weights on these two edges to be just $r$. We justify this heuristics and the asymmetry in the AE is that (a) we find such a crafted decoder to be comparable or complementary to feature engineering in the AE's input, and (b) we anticipate that if we add  more neurons in layer 4, and/or more layers, with diverse activation functions, the PoaC training will be able to approximate  the $\cos()$
and  $\sin()$ functions, and/or learn the weights that we have fixed. 

After identifying a working architecture for PoaC AEs, 
we prepared a dataset of such AEs by training each one with a set of 200  points. In most cases, the class AEs converged nicely. 

To train the PoaC MAE we created training and test sets of AEs for circles with radiuses in fixed intervals in the range $1 \leq r \leq 10$.  

The MAE has 9 layers with 35-20-10-4-1-4-10-20-35 neurons respectively, and ReLU and linear activations. 
The number 35 is the total number of trainable weights and biases in the PoaC 2-8-1-2-2 AEs, while other weights and biases are fixed, as described in the heuristics above.  
For each circle we created multiple AEs, using a different set of 200 random points in each AE. 

The training of an MAE for circles did not readily converge so we took the following heuristic steps. First, we narrowed down the example from Points-on-a-Circle to Points-on-Arc (PoA). The individual class AEs were generated as for circles, but instead of the range of points being chosen in angles from $-\pi$ to $\pi$, they were between $\pi/6$ and  $\pi/3$. 

We then normalized the AEs as follows. Different AEs for a given circle may actually be isomorphic to each other, and similar AEs may look very different due to the order of neurons. Hence, we sorted the 8 neurons in layer 2 of the PoaL AEs by the product of the two weights on the edges from the two neurons in layer 1 to the same neuron in layer 2. As in PoaL, in training the PoA MAE we employed execution-driven loss calculation, using the 35 output values to populate the weights and biases in a skeleton 2-8-1-2-2 AE, with the above-specified activations. 
The MAE training was successful and the resulting model is available in (URL to be provided in  CR version).  

This example can be extended too; e.g.,  narrow down the family to classes of arcs for which trained AEs perform well, explore simpler MAE models, perhaps with different activation functions, and interpret the codes of the AEs and MAE for insights that may accelerate training and simplify models.  

\section{Related work}\label{sec:relwork}

The construction of AEs and MAEs aims to discover relationship patterns between properties of individuals and groups within an evolving population. Such goals may also be pursued analytically, using classical statistical methods, including principal component analysis (PCA)~\cite{abdi2010PCA} and meta-PCA~\cite{kim2018metaPCA}. 

Since our motivation lies in modeling biological processes --- often depending on reproduction and sustainment --- and as autoencoding can be viewed also as an imperfect reproduction mechanism, we place AEs at the center of our approach. Other important modeling aspects, such as precision, efficiency, succinctness, and formal provability can then be addressed within this biologically oriented framework. 
Certain limitations of AEs and MAEs, as compared with analytical methods, may be sidestepped in natural settings. For example, the real-world data may lie within a narrow, computationally safe band, rarely reaching theoretical worst-case scenarios.

The pursuit of autoencoding in multi-class environments and contexts is not new ~\cite{deng2021FeatureDisentanglementAE,liu2019compactFeatureLearning,wangBreckon2023DomainAdaptationAE,abinaya2023cascadingAEMultiClass}.  However, the projects we have seen are application-specific, often in image processing, where the classes being autoencoded and the associated challenges are pre-determined. For example, the goal may be  to develop an AE-based technique for domain adaptation --- transferring learned knowledge between image domains that differ in quality or context.

By contrast, our broad interest is in: (a) classes whose contents are not known in advance, as they evolve dynamically, but whose differences are confined, as they evolve from each other or from common ancestors; and (b) identifying a domain independent framework or computational principles, that can support a theory of universal autoencoding mechanisms.  

In addition, the focus on modeling nature may uncover  ways to combine  simplicity with robustness that may be inadequate in a more general machine learning setting: As natural autoencoding is carried out in parallel by multiple instances, if some individuals do not survive due to autoencoding-related failures, others may continue executing the same imperfect process for other cases, yielding acceptable population-wide results.

\section{Conclusion and Future Research}\label{sec:conclusion}
We have introduced the concept of meta-autoencoders for families of autoencodable classes as a direction for a modular domain-independent mechanism for  parameterizing the differences among collections of  classes. We view future research of meta-autoencoding in two trajectories.  

The first is incorporating AEs and MAEs in modeling biological evolution as described in Section~\ref{sec:intro}, including reproduction (both sexual and asexual), recurrence (manifested in feeding back the outputs of AEs into fresh training cycles), clustering within populations to delineate species, and parallel autoencoding of interacting species. In this context we plan to also study \textbf{distributional autoencoding},
where each reconstructed point is allowed to differ significantly from the corresponding input point, which might model its parent, while maintaining the input property distribution in the reconstructed population. Such distributional autoencoding differs from variational autoencoding, where stochasticity is in the latent layer rather than in the output population. 
 
The second research direction is the theory of MAEs, including: extending the constructive definition offered here,  demonstrating MAE construction in additional domains, multi-level meta-autoencoding, i.e., an MAE for a collection of MAEs, studying categories of families of classes for which the  existence or non-existence of MAEs may be proven a-priori, interpreting the code of AEs and MAEs as well as the relations
between multiple AEs for the same class and across classes; and, (v)~developing methods and tools for systematic construction of meta-autoencoders in diverse applications. 

\subsubsection{\ackname} 
This research was funded in part by an NSFC-ISF grant to DH,  issued jointly by the  National Natural Science Foundation of China (NSFC) and the Israel Science Foundation (ISF grant 3698/21). Additional support was provided by a research grant to DH from Louis J. Lavigne and Nancy Rothman, the Carter Chapman Shreve Family Foundation, Dr. and Mrs. Donald Rivin, and the Estate of Smigel Trust.
%

\bibliographystyle{splncs04}






\end{document}